\documentclass[sigconf,screen]{acmart}
\AtBeginDocument{%
  }

\setcopyright{acmlicensed}
\copyrightyear{2025}
\acmYear{2025}
\acmDOI{XXXXXXX.XXXXXXX}
\acmConference[MM '25]{ACM Multimedia 2025}{October 27-31, 2025}{Dublin, Ireland}
\acmBooktitle{Proceedings of the 33rd ACM International Conference on Multimedia (MM '25), Octobe 27--31, 2025, Dublin, Ireland}
\acmISBN{978-1-4503-XXXX-X/2018/06}

\usepackage{url}
\usepackage[T1]{fontenc}
\usepackage{graphicx}
\usepackage{booktabs}
\usepackage{threeparttable}
\usepackage{multirow} 
\usepackage{longtable}
\usepackage{diagbox}
\usepackage{multicol}
\usepackage{makecell}
\usepackage{algorithm, algorithmic}
\usepackage{setspace}
\usepackage{subcaption}

\usepackage{listings}
\usepackage{xcolor}
\usepackage{pgffor}
\usepackage{wrapfig}
\usepackage{tikzsymbols}
\acmSubmissionID{174}

\begin{document}


\title{ICAS: Detecting Training Data from Autoregressive Image Generative Models}

\author{Hongyao Yu}
\email{chrisqcwx@gmail.com}
\affiliation{%
  \institution{Tsinghua Shenzhen International Graduate School, Tsinghua University\\Harbin Institute of Technology, Shenzhen}
  \city{Shenzhen}
  \state{Guangdong}
  \country{China}
}

\author{Yixiang Qiu}
\email{qiu-yx24@mails.tsinghua.edu.cn}
\affiliation{%
  \institution{Tsinghua Shenzhen International Graduate School, Tsinghua University}
  \city{Shenzhen}
  \state{Guangdong}
  \country{China}
}

\author{Yiheng Yang}
\email{2023311d15@stu.hit.edu.cn}
\affiliation{%
  \institution{Harbin Institute of Technology, Shenzhen}
  \city{Shenzhen}
  \state{Guangdong}
  \country{China}
}

\author{Hao Fang}
\email{fang-h23@mails.tsinghua.edu.cn}
\affiliation{%
  \institution{Tsinghua Shenzhen International Graduate School, Tsinghua University}
  \city{Shenzhen}
  \state{Guangdong}
  \country{China}
}

\author{Tianqu Zhuang}
\email{zhuangtq23@mails.tsinghua.edu.cn}
\affiliation{%
  \institution{Tsinghua Shenzhen International Graduate School, Tsinghua University}
  \city{Shenzhen}
  \state{Guangdong}
  \country{China}
}

\author{Jiaxin Hong}
\email{acanghong425@gmail.com}
\affiliation{%
  \institution{Harbin Institute of Technology, Shenzhen}
  \city{Shenzhen}
  \state{Guangdong}
  \country{China}
}

\author{Bin Chen}
\authornote{Corresponding author.}
\email{chenbin2021@hit.edu.cn}
\affiliation{%
  \institution{Harbin Institute of Technology, Shenzhen}
  \city{Shenzhen}
  \state{Guangdong}
  \country{China}
}

\author{Hao Wu}
\email{wu-h22@mails.tsinghua.edu.cn}
\affiliation{%
  \institution{Tsinghua Shenzhen International Graduate School, Tsinghua University\\Shenzhen ShenNong Information Technology Co., Ltd.}
  \city{Shenzhen}
  \state{Guangdong}
  \country{China}
}

\author{Shu-Tao Xia}
\email{xiast@sz.tsinghua.edu.cn}
\affiliation{%
  \institution{Tsinghua Shenzhen International Graduate School, Tsinghua University}
  \city{Shenzhen}
  \state{Guangdong}
  \country{China}
}







\renewcommand{\shortauthors}{Yu et al.}


\def\ie{\textit{i.e.}}
\def\eg{\textit{e.g.}}
\newcommand\todo[1]{{\color{blue} (TODO: #1)}}

\begin{abstract}

Autoregressive image generation has witnessed rapid advancements, with prominent models such as scale-wise visual auto-regression pushing the boundaries of visual synthesis. However, these developments also raise significant concerns regarding data privacy and copyright.
%
In response, training data detection has emerged as a critical task for identifying unauthorized data usage in model training. To better understand the vulnerability of autoregressive image generative models to such detection, we conduct the first study that applies membership inference to this domain. 
%
%
Our approach comprises two key components: implicit classification and an adaptive score aggregation strategy. First, we compute the implicit token-wise classification score within the query image. 
Then we propose an adaptive score aggregation strategy to acquire a final score, which places greater emphasis on the tokens with lower scores. 
A higher final score indicates that the sample is more likely to be involved in the training set. 
Extensive experiments demonstrate the superiority of our method over those designed for LLMs, in both class-conditional and text-to-image scenarios.
Moreover, our approach exhibits strong robustness and generalization under various data transformations. 
Furthermore, sufficient experiments suggest two novel key findings: (1) A linear scaling law on membership inference, exposing the vulnerability of large foundation models. (2) Training data from scale-wise visual autoregressive models is easier to detect than other autoregressive paradigms.
Our code is available at \url{https://github.com/Chrisqcwx/ImageAR-MIA}.
  
\end{abstract}



\received{20 February 2007}
\received[revised]{12 March 2009}
\received[accepted]{5 June 2009}


\maketitle

\section{Introduction}

\begin{figure}[!htbp]
    \begin{subfigure}[b]{\linewidth}
    \begin{minipage}[b]{\textwidth}
                \centering
        \includegraphics[width=\textwidth]{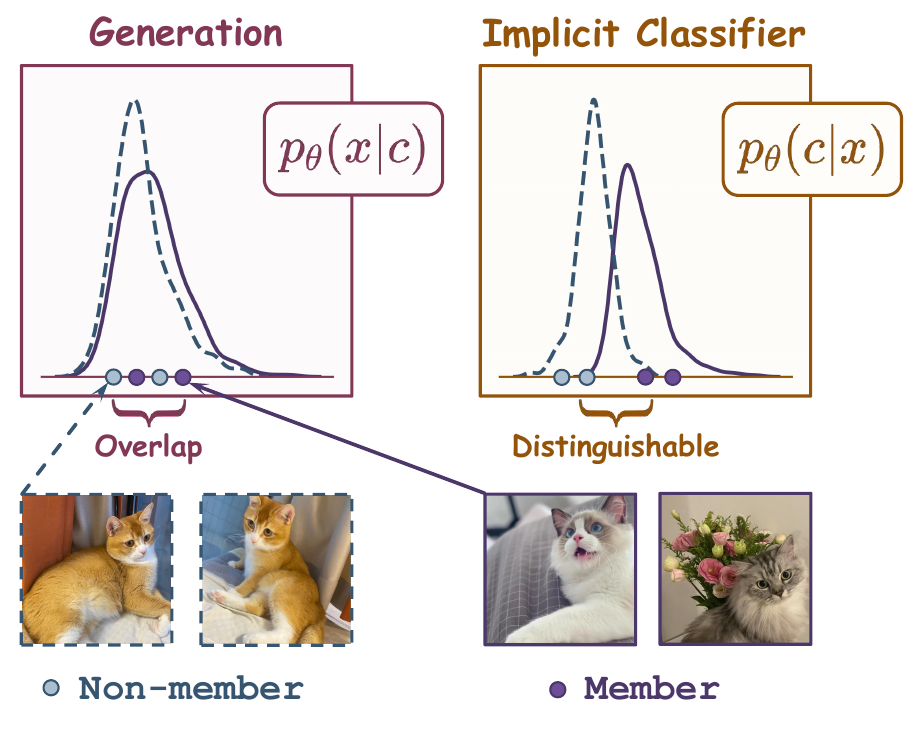}
    \end{minipage}%
    \end{subfigure}
    \caption{MI methods developed for LLMs, which leverage the generative probability, fail to distinguish the distribution shift between member and non-member samples. In contrast, the implicit classifier can distinguish them.}
    \label{fig: intro}
\end{figure}

In the past few years, autoregressive image generation has advanced rapidly, with scale-wise visual autoregression (VAR) \cite{var} emerging as a prominent representative. VAR not only achieves superior image quality compared to diffusion models but also accelerates the image generation process, establishing it as a mainstream paradigm in image generation, alongside GANs and diffusion models \cite{var, varclip, cca}.
However, the impressive generative capabilities of visual autoregressive models rely heavily on large-scale image datasets, primarily sourced from the web \cite{mmmia,clid}. This practice poses potential risks to privacy \cite{ifgmi,miasurvey,mibench} and security \cite{calor,ssd,gifd}.
%
Membership inference (MI) \cite{mink,secmi,mmmia} has become a widely adopted technique for detecting unauthorized data usage during model training. This enables data owners to identify whether their data has been used in the training process without permission, which can lead to potential misuse or legal violations.
%

To fully understand the robustness of such models to protect their training data, we propose the first membership inference research specifically targeting VARs. Although this visual autoregressive paradigm shares similarities to autoregressive large language models (LLMs), the membership inference methods developed for LLMs \cite{mink,minkpp,renyi} fail to perform effectively on autoregressive image models due to their strong generalization capabilities, as demonstrated in the left part of Figure \ref{fig: intro}.

%
To address this challenge, we introduce a novel membership inference method ICAS tailored for autoregressive image generative models. Unlike detection methods for LLMs that estimate the generation probability for a given condition, our approach considers the probability of the condition corresponding to the image, referred to as the implicit classification, which can effectively distinguish between member and non-member samples, as shown in Figure \ref{fig: intro}.
Building upon this, we propose a membership inference method based on the implicit classifier (IC). Specifically, we fist compute the implicit classification score of the image tokens in the query image. Next, we employ an adaptive score aggregation strategy (AS) to combine these token-level scores into a final score, which focuses more on low-score tokens. A higher final score indicates a greater likelihood that the image is included in the training data. 

We conduct extensive experiments on various visual autoregressive models in different scenarios, including class-conditional and text-to-image generation. For comparison, we also implement detection methods designed for large language models, due to the similarities in the autoregressive paradigm. Our experimental results demonstrate that our proposed method outperforms existing detection methods for LLMsin detecting the training data of visual autoregressive models. Additionally, robustness analysis and ablation studies confirm the robustness and efficacy of our approach. We also explore the relationship between model size and detection performance, revealing a linear scaling law for membership inference in relation to model parameters, indicating that larger models are more vulnerable to membership inference. Moreover, we find that the detection from the scale-wise autoregressive model is much easier than other autoregressive paradigms.

We summarize our contributions as follows: 
\begin{itemize}
    \item We are the first to explore membership inference for training data detection in autoregressive image generative models. Our method contains an implicit classifier and an adaptive score aggregation strategy.
    \item Experimental results reveal that our methods outperform competitive baselines and demonstrate the robustness and generalizability of our method across various scenarios.
    \item We explore a linear scaling law for membership inference in autoregressive image generative models, and discover it is easier to detect training data on large or scale-wise visual autoregressive models.
\end{itemize}


\section{Related Work}
\subsection{Membership Inference on Diffusion Models.}

Recently, diffusion models have demonstrated superior performance in generative tasks \cite{ldm,sd3,kolor}. To ensure that image data is not used without authorization, some membership inference \cite{naiveloss,pfami} algorithms have been proposed to detect whether images are used for model training. Naive Loss \cite{naiveloss} attack is the first proposed to leverage the prediction loss on the random Gaussian loss to detect the training sample on diffusion models, which is implemented by comparing the distance between the Gaussian noise and the predicted noise. Proximal Initialization Attack (PIA) \cite{pia} advances the Naive Loss \cite{naiveloss} by utilizing the predicted noise in the clean image instead of the random noise. 
In addition to using only the sample itself for evaluation, PFAMI \cite{pfami} introduces the neighbor samples, acquired through image transformations. The research finds higher denoising loss gap between the origin sample and neighbor samples indicates higher likelihood for the member samples.
Instead of detect from the original clean image, \cite{secmi} proposes the SecMI method. They adopt the DDIM inversion technique to reverse the image to timestep $t$, and calculate the approximated posterior estimation error. Similarly, \cite{mmmia} also leverages the DDIM inversion result, they compute the SSIM metric between it and the clean image. \cite{clid} further identify the conditional overfitting phenomena, they compare the predicted noise with and without the corresponding condition for membership inference. However, although membership inference for image generation in diffusion models has been extensively studied, it cannot be directly transferred to autoregressive image generation, owing to its completely different generation paradigm.

\subsection{Membership Inference on LLMs} 

In addition to image generative models, membership inference on large language models (LLMs) similarly attracts extensive attention from the research community. 
For instance, the Min-$k\%$ method \cite{mink} assumes that the member samples are less likely to contain outlier tokens with low log probabilities. They identify the $k\%$ tokens with the lowest log probabilities, averaging them to detect member samples. 
Building upon this foundational work, \cite{minkpp} further emphasizes the local maxima issue, considering that the ground-truth token is more likely to have a larger log probability than other tokens in a member sample. Instead of directly using the log probability, they subtract the average token probability score from the log probability of each token and normalize it
In contrast to methods based on the average probability of the lowest $k\%$ tokens, DC-PDD \cite{dcpdd} leverages the divergence between the token probability distribution within the text and the token frequency distribution across the reference corpus to identify training data, which mitigates misclassification issues that arise when non-member texts contain numerous common tokens with high predicted probabilities.
Instead of using the predicted probability, \cite{infillingscore} proposes to calculate the infilling log likelihood for tokens. This method not only considers the preceding tokens but also incorporates subsequent ones for a more comprehensive and accurate analysis.
Lastly, \cite{FineTuning} explores the behavior gap in fine-tuning between member and non-member samples. Their findings show that the gap in log probability before and after fine-tuning is smaller for member samples compared to non-members. 
%

\subsection{Autoregressive Image Generative Models.}

In recent years, the field of autoregressive image generation has developed rapidly, and a variety of autoregressive image generation paradigms have emerged. VQVAE \cite{vqvae} first demonstrates the ability to decompose a 2D image into 1D discrete token sequences. VQGAN \cite{VQGAN} upgrade from VQVAE by introducing the perceptual and discriminator loss, as well as a raster-scan autoregressive transformer to generate image tokens. LlamaGen proposes to use the Llama transformer architecture to execute the raster-scan process, combining VQVAE and exceeding the stable diffusion model \cite{ldm}, and PAR \cite{par} accelerates it by parallel decoding. MaskGIT \cite{maskgit} proposed a mask-based method, using a bidirectional transformer to predict the mask tokens. Beyond 1D raster scanning, NAR \cite{nar} predicts the neighbor tokens in all directions.

Among the many autoregressive generation paradigms, the scale-wise visual autoregressive model (VAR) \cite{var} is an outstanding representative. \cite{var} first proposes this generative paradigm. It leverages a multi-scale image tokenizer to transfer an image into image tokens with different scales. The transformer predicts the next-scale tokens with previous-scale tokens. There are many follow-up improvements based on this foundation work. VAR-CLIP \cite{varclip} replace the class condition with CLIP embedding, extending it from class-condition generation to text-to-image generation. M-VAR \cite{mvar} upgrade it with a mamba architecture \cite{mamba}. CCA \cite{cca} considers the classifier-free guidance into training to avoid the extra cost for the guidance in the inference time. The rapid development and widespread adoption of this technology highlight the urgent need to study its privacy and security issues.
\section{Method}

\subsection{Preliminaries}

\noindent\textbf{Membership Inference.} 
Let $\mathbb{D}$ be a dataset drawn from the real data distribution. The model owner uses a subset $\mathbb{D}_{mem}\subseteq \mathbb{D}$ to train a model. The remaining part is denoted as the hold-out dataset $\mathbb{D}_{out}=\mathbb{D}\backslash\mathbb{D}_{mem}$. 
In the context of conditional image generation models, each data point consists of an image $\mathbf{x}$ and a corresponding condition $c$. The goal of a membership inference (MI) is to determine whether a given data point $(\mathbf{x}, {c})$ is used in the training of the image generator. The membership inference can mainly be divided into reference-based methods \cite{watson2021importance,carlini2022membership,shi2024learning} and reference-free methods \cite{mink,minkpp,renyi}.
Reference-based methods typically require the training of shadow models on similar distributions. However, these approaches depend on prior knowledge of the training data, and the cost of training additional models makes them impractical for large-scale models \cite{mink}. Consequently, we focus on reference-free methods that do not require prior knowledge of the training data or the details of the training process. The membership inference algorithm for conditional image generation can be formulated as:
\begin{equation}
    \mathcal{M}(\mathbf{x}, {c}) = \mathbf{1}[\mathrm{Score}(\mathbf{x},  {c})\ge\tau],
\end{equation}
where $\mathcal{M}(\mathbf{x},  {c})=1$ means the data point $(\mathbf{x},  {c})$ is used to train the target model, $\tau$ is a threshold, and $\mathbf{1}[\cdot]$ is an indicator function.

\noindent\textbf{Scale-wise Visual Autoregressive Models.} 
\label{sec:var preliminary}
Scale-wise Visual Autoregressive Model (VAR) \cite{var} contains a multi-scale VQVAE and a visual autoregressive transformer.

The multi-scale VQVAE is used to tokenize the image into a series of discrete image token maps with $K$ different scale levels. It includes an encoder $\mathcal{E}$, a decoder $\mathcal{D}$, and a quantizer $\mathcal{Q}$ with a codebook $Z\in \mathbb{R}^{V\times C}$, where $V$ is the vocabulary size and $C$ is the channel dimension. 
The encoder tokenizes the image $\mathbf{x}$ into a feature map $f_K$:
\begin{equation}
    f_K=\mathcal{E}(\mathbf{x})\in \mathbb{R}^{C\times h_K\times w_K},
\end{equation}
where $h_K\times w_K$ is the latent resolution of the feature map. Then the feature map is quantized into $K$ multi-scale token maps $R=(r_1, r_2,\dots,r_K)$. The $r_k$ contains $h_k\times w_k$ tokens, and the number increases when the scale level $k$ gets larger. The $k$-th residual map is defined as $f'_k=Z(r_k)$. The $k$-th feature map $f_k$ is the cumulative sum of the residual maps:
\begin{equation}
    f_k = \sum_{i=1}^k {\mathrm{Upsample}(f_i', (h_k\times w_k))},
\end{equation}
where $\mathrm{Upsample}(f', (h\times w))$ means to resize the residual map $f'$ into size $h\times w$ with bilinear upsampling function. The reconstructed image can be decoded with the last feature map:
\begin{equation}
    \mathbf{x}' = \mathcal{D}(f_K).
\end{equation}

The visual autoregressive transformer ${T}_\theta$ parameterized with weight $\theta$ is used to predict the next-scale token maps using tokens with lower scale levels:
\begin{equation}
\label{eq:predict}
    p_\theta(r_k|c, r_1,\dots,r_{k-1})={T}_\theta(c, f_{k-1}^{\mathrm{up}})
\end{equation}
\begin{equation}
    f_{k-1}^{\mathrm{up}}=\mathrm{Upsample}(f_{k-1},  (h_k\times w_k)),
\end{equation}
where $c$ is the condition. 
The recursive formula of the feature map can also be written as:
\begin{equation}
    f_k = Z(r_{k}) + \mathrm{Upsample}(f_{k-1}, (h_k\times w_k)),
\end{equation}
where $r_k$ is sampled from $p_\theta(r_k|c, r_1,\dots,r_{k-1})$. In this paper, we focus on the membership inference on the visual autoregressive transformer.

\begin{figure*}[!tbp]
    \begin{subfigure}[b]{\linewidth}
    \begin{minipage}[b]{\textwidth}
                \centering
        \includegraphics[width=\textwidth]{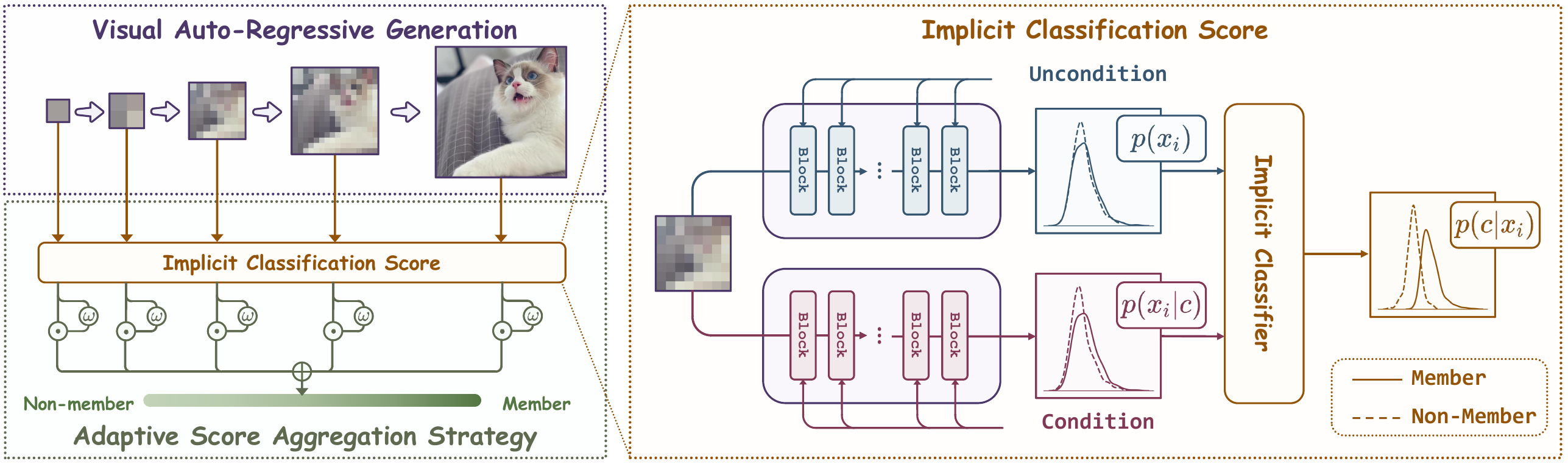}
    \end{minipage}%
    \end{subfigure}
    \caption{Overview of our membership inference method. The distribution is based on the experimental results of VAR-$d24$.}
    \label{fig: main method}
\end{figure*}

\subsection{Detection via Implicit Classifier}

As outlined in Sec. \ref{sec:var preliminary}, an image $\mathbf{x}$ can be decomposed into $K$ token maps $(r_1,r_2,\dots,r_K)$. Consequently, the likelihood of the image can be expressed as:
\begin{equation}
\begin{aligned}
    p_\theta(\mathbf{x}|c) &= \prod_{k=1}^K p_\theta(r_k|c, r_1,\dots, r_{k-1}) \\
    &=\prod_{k=1}^K \prod_{i=1}^{h_k}\prod_{j=1}^{w_k} p_\theta(r_k^{(i, j)}|c, r_1,\dots, r_{k-1}),
\end{aligned}
\end{equation}
where $r_k^{(i, j)}$ represents the token with coordinates $(i, j)$ on $r_k$. For simplicity, we expand all tokens into a sequence and ignore the conditions of the preceding token maps for simplicity so that the probability can be written as:
\begin{equation}
    p_\theta(\mathbf{x}|c)=\prod_{i=1}^N p_\theta(x_i|c),
\end{equation}
where $N=\sum_{k=1}^K h_k\times w_k$ is the total number of tokens.

To detect whether an image $\mathbf{x}$ belongs to the training set, recent studies exploit an overfitting phenomenon. The overfitting issue manifests a higher prediction probability for the ground-truth tokens in the member set \cite{mink,minkpp}.
However, this assumption fails when it comes to the visual autoregressive models, which exhibit strong generalization ability. As shown in the right portion of Figure \ref{fig: main method}, the distributions of log-likelihoods for tokens (\ie, $p_\theta(x_i|c)$) from member samples and non-member samples are similar, making it challenging to differentiate between them.

\begin{figure}[!htbp]
    \begin{subfigure}[b]{\linewidth}
    \begin{minipage}[b]{\textwidth}
                \centering
        \includegraphics[width=\textwidth]{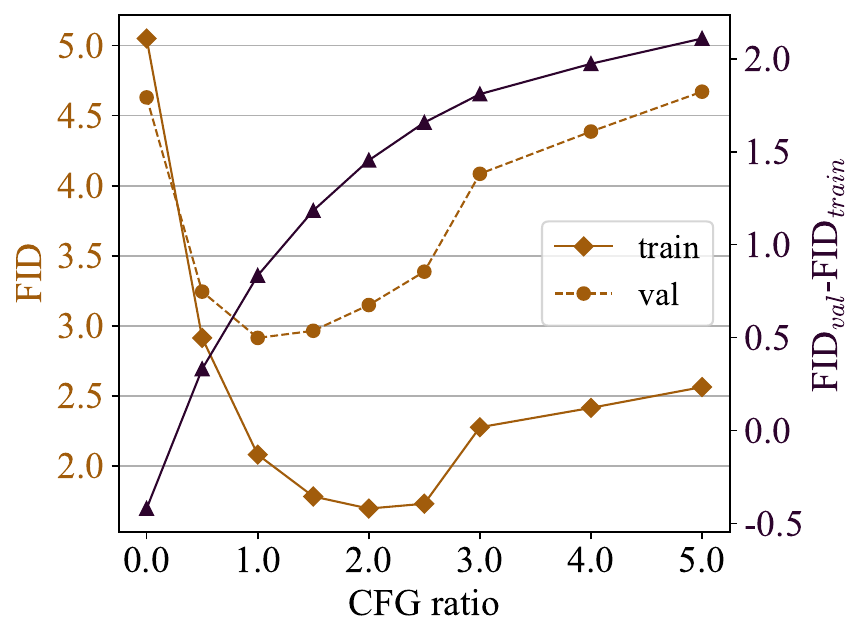}
    \end{minipage}%
    \end{subfigure}
    \caption{FID evaluation results with different CFG ratios on training set and validation set.}
    \label{fig: motivation fid}
\end{figure}

Classifier-free guidance (CFG), as proposed by \cite{cfg}, involves comparing the model predictions with and without a specific condition, such as a class or a text condition. It utilizes an implicit classifier $p_\theta(c|\mathbf{x})$ to guide the generation, \ie, $\tilde{p}_\theta(\mathbf{x}|c)=p_\theta(\mathbf{x}|c)p^\omega_\theta(c|\mathbf{x})$, where $\theta$ is the parameters of the generative model, and $\omega$ is the CFG ratio. The implicit classifier is implemented as follows:
\begin{equation}
    p_\theta(c|\mathbf{x})\propto \frac{p_\theta(\mathbf{x}|c)}{p_\theta(\mathbf{x})}.
\end{equation}
The CFG guidance works by adjusting the generation process to push the result in the direction of high confidence for the target condition while avoiding regions that correspond to other classes. In the practice of the generation using diffusion models, the CFG ratio is typically set to a value like $7.5\gg 1$, underscoring its significant role in conditional image generation.
To validate this effect in visual autoregressive models, we use the VAR-$d30$ \cite{var} trained on the training set of ImageNet \cite{imagenet} to generate $50000$ images with a series of different CFG ratios. 
%
Then we calculate the FID score \cite{fid} of the generated images. We use two reference datasets to calculate FID, the training set (member) and the validation set (non-member), and the evaluation results are denoted as FID$_{train}$ and FID$_{val}$. 
The evaluation results presented in Figure \ref{fig: motivation fid} shows that a higher CFG ratio leads to a higher gap between FID$_{train}$ and FID$_{val}$, which means that the distribution of the generated images is closer to the training set with the help of the classifier-free guidance. This finding further highlights a strong potential from the implicit classifier in distinguishing member samples from non-member samples.

Inspired by these findings, we propose an implicit classifier-based detection method. It leverages the difference between conditional and unconditional predicted probabilities to determine whether a sample is a member. Specifically, the method approximates the implicit classification score by:
\begin{equation}
\begin{aligned}
  \log p_\theta (c|\mathbf{x}) &\propto \log \frac{p_\theta(\mathbf{x}|c)}{p_\theta(\mathbf{x})}\\
    &=\log\frac{\prod_{i=1}^N p_\theta(x_i|c)}{\prod_{i=1}^N p_\theta(x_i)} \\
    &= \sum_{i=1}^N (\log  p_\theta(x_i|c)-\log p_\theta(x_i)) \\
    &\overset{\text{def.}}{=} \mathrm{Score}(\mathbf{x}, c).
\end{aligned}
\end{equation}
Intuitively, the score is the accumulation of the difference between the conditional and unconditional log probability of each token, and the implicit classification score of each image token can be formulated as follows:
\begin{equation}
    \mathrm{Score}(x_i,c)=\log  p_\theta(x_i|c)-\log p_\theta(x_i).
\end{equation}

The right portion of Figure \ref{fig: main method} demonstrates the effectiveness of the implicit classifier (\ie, $p_\theta(c|x_i)$). The distribution of log probability with or without condition is similar between member samples and non-member samples, resulting in a low Area-Under-the-ROC-curve (AUROC). However, when we compute the implicit classification score, a clear distinction emerges between member and non-member samples, significantly improving the model's ability to distinguish between them. This demonstrates the power of our implicit classifier-based method in enhancing the detection performance of membership in visual autoregressive models.

\subsection{Adaptive Score Aggregation Strategy}

\begin{figure}[!htbp]
    \begin{subfigure}[b]{.7\linewidth}
    \begin{minipage}[b]{\textwidth}
                \centering
        \includegraphics[width=\textwidth]{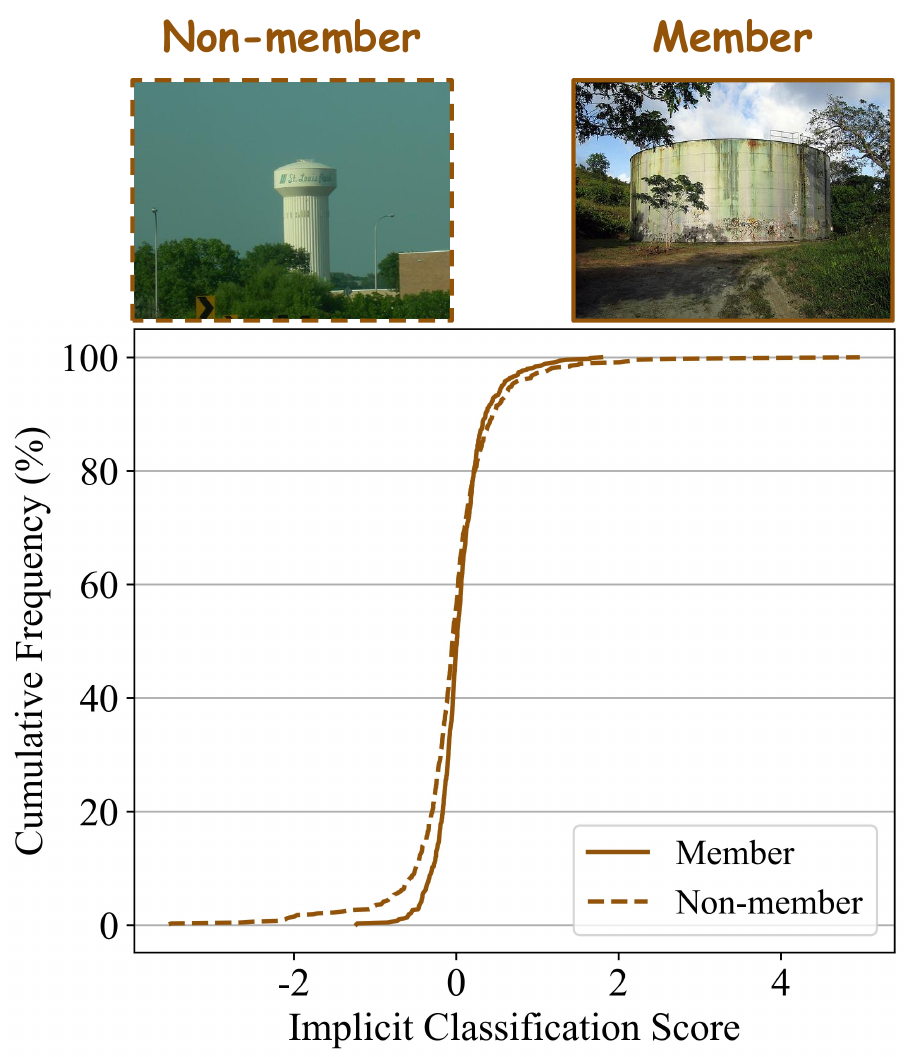}
    \end{minipage}%
    \end{subfigure}
    \caption{Cumulative distribution of token scores from a specific member and a non-member sample. A few tokens in the non-member sample have high scores, while more have low scores.}
    \label{fig:individual}
\end{figure}

In the previous section, we defined the implicit classification score of each image token, and the overall score of the image as the sum of all token scores.
%
However, the difficulty of predicting tokens for different images varies significantly. It is possible for a few tokens from non-member samples to receive high scores from the implicit classifier, with an example from Figure \ref{fig:individual}. These high-scoring tokens, though limited in number, can negatively impact the performance of membership inference. On the other hand, due to the overfitting phenomenon during training, the probability of low token scores in member samples tends to be smaller, while non-member samples generally have more tokens with low scores \cite{mink}. 
To address this issue and better distinguish between member and non-member samples, it is essential to assign more weight to tokens with lower scores and less weight to those with higher scores.
To achieve this, we introduce a adaptive score aggregation approach, which puts more emphasis on the tokens with lower scores:
%
\begin{equation}
\begin{aligned}
    \mathrm{Score}(\mathbf{x},c) &= \sum_{i=1}^N \omega_i\cdot\mathrm{Score}(x_i,c),
\end{aligned}
\end{equation}
where $\omega_i>0$ is the weight of the $i$-th token, and it should be larger when the $\mathrm{Score}(x_i,c)$ comes lower.
In practice, we define the weight $\omega_i$ as follows:
\begin{equation}
    \omega_i=\frac{1}{a+\exp(b\cdot \mathrm{Score}(x_i, c))},
\end{equation}
where $a>0$ and $b>0$ are hyperparameters. Therefore, the score of the total image can be formulated as:
\begin{equation}
    \mathrm{Score}(\mathbf{x},c) =\sum_{i=1}^N \frac{1}{a+
    (\frac{p_\theta(x_i|c)}{p_\theta(x_i)})^b
    }\log \frac{p_\theta(x_i|c)}{p_\theta(x_i)}.
\end{equation}

\begin{table*}[!ht]
    \setlength{\tabcolsep}{5pt}
    \normalsize
    \centering
    \caption{Experiment results in VAR models}
    \label{tab:main var}
    \begin{threeparttable} 
    \resizebox{.95\linewidth}{!}{
    \begin{tabular}{ccccccccccccc}
        \toprule
 \multirow{2}{*}{{Method}} 
 & \multicolumn{3}{c}{VAR-$d16$} & \multicolumn{3}{c}{VAR-$d20$}& \multicolumn{3}{c}{VAR-$d24$}& \multicolumn{3}{c}{VAR-$d30$} \\

 \cmidrule(lr){2-4}\cmidrule(lr){5-7}\cmidrule(lr){8-10}\cmidrule(lr){11-13} & $\uparrow$ {AUROC} & $\uparrow$ {TPR@5\%} & $\uparrow$ ASR  & $\uparrow$ {AUROC} & $\uparrow$ {TPR@5\%} & $\uparrow$ ASR  & $\uparrow$ {AUROC} & $\uparrow$ {TPR@5\%} & $\uparrow$ ASR  & $\uparrow$ {AUROC} & $\uparrow$ {TPR@5\%} & $\uparrow$ ASR \\ \midrule

$Loss$ & 0.5223 & 0.0580 & 0.5163 & 0.5468 & 0.0707 & 0.5332 & 0.6027 & 0.1058 & 0.5752 & 0.7754 & 0.3084 & 0.7090  \\
$Min$-$k\%$  & 0.5366 & 0.0653 & 0.5244 & 0.5813 & 0.0798 & 0.5570 & 0.6648 & 0.1261 & 0.6186 & 0.8593 & 0.4095 & 0.7803  \\
$Min$-$k\%$++  & 0.5213 & 0.0572 & 0.5140 & 0.5402 & 0.0659 & 0.5267 & 0.5941 & 0.0938 & 0.5676 & 0.7778 & 0.2919 & 0.7039  \\
\textit{R\'enyi} & 0.5315 & 0.0660 & 0.5211 & 0.5577 & 0.0784 & 0.5412 & 0.6332 & 0.1251 & 0.5948 & 0.8520 & 0.4597 & 0.7681  \\
{$ICAS (ours)$} & \textbf{0.6838} & \textbf{0.1417} & \textbf{0.6345} & \textbf{0.8402} & \textbf{0.3494} & \textbf{0.7628} & \textbf{0.9624} & \textbf{0.7824} & \textbf{0.9001} & \textbf{0.9990} & \textbf{0.9997} & \textbf{0.9897}  \\

\bottomrule
    \end{tabular}
    }
    \end{threeparttable}
    
\end{table*}





\begin{table*}[!ht]
    \setlength{\tabcolsep}{5pt}
    \normalsize
    \centering
    \caption{Experiment results in VAR-CCA models}
    \label{tab:main cca}
    \begin{threeparttable} 
    \resizebox{.95\linewidth}{!}{
    \begin{tabular}{ccccccccccccc}
        \toprule
 \multirow{2}{*}{{Method}} 
 & \multicolumn{3}{c}{VAR-CCA-$d16$} & \multicolumn{3}{c}{VAR-CCA-$d20$}& \multicolumn{3}{c}{VAR-CCA-$d24$}& \multicolumn{3}{c}{VAR-CCA-$d30$} \\

 \cmidrule(lr){2-4}\cmidrule(lr){5-7}\cmidrule(lr){8-10}\cmidrule(lr){11-13} & $\uparrow$ {AUROC} & $\uparrow$ {TPR@5\%} & $\uparrow$ ASR  & $\uparrow$ {AUROC} & $\uparrow$ {TPR@5\%} & $\uparrow$ ASR  & $\uparrow$ {AUROC} & $\uparrow$ {TPR@5\%} & $\uparrow$ ASR  & $\uparrow$ {AUROC} & $\uparrow$ {TPR@5\%} & $\uparrow$ ASR \\ \midrule

$Loss$ & 0.5310 & 0.0689 & 0.5217 & 0.5702 & 0.1025 & 0.5493 & 0.6737 & 0.2306 & 0.6266 & 0.9177 & 0.7483 & 0.8516  \\
$Min$-$k\%$  & 0.5543 & 0.0697 & 0.5382 & 0.6321 & 0.0934 & 0.5947 & 0.7908 & 0.1918 & 0.7257 & 0.9774 & 0.8751 & 0.9427  \\
$Min$-$k\%$++  & 0.5135 & 0.0504 & 0.510 & 0.5148 & 0.0485 & 0.5137 & 0.5074 & 0.0646 & 0.5109 & 0.6039 & 0.1333 & 0.5871  \\
\textit{R\'enyi} & 0.5180 & 0.0609 & 0.5120 & 0.5407 & 0.0782 & 0.5272 & 0.6022 & 0.1382 & 0.5717 & 0.8326 & 0.5369 & 0.7607  \\
$ICAS (ours)$  & \textbf{0.6032} & \textbf{0.1016} & \textbf{0.5780} & \textbf{0.7642} & \textbf{0.2463} & \textbf{0.7049} & \textbf{0.9279} & \textbf{0.6605} & \textbf{0.8583} & \textbf{0.9980} & \textbf{0.9969} & \textbf{0.9869}  \\
 
\bottomrule
    \end{tabular}
    }
    \end{threeparttable}
    
\end{table*}

{

\begin{figure*}[!htbp]
    \begin{minipage}[b]{\textwidth}
    \begin{minipage}[b]{0.245\textwidth}
                \centering
        \includegraphics[width=\textwidth]{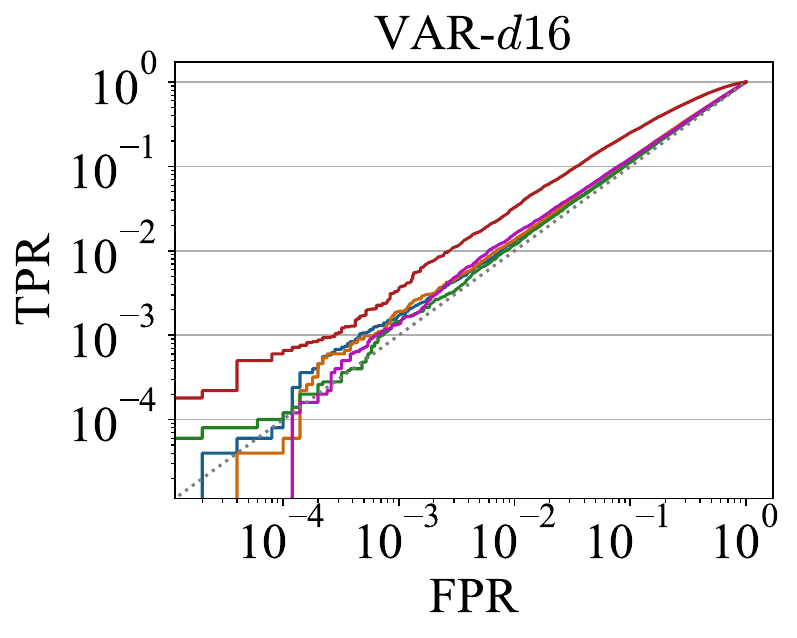}
    \end{minipage}%
    \begin{minipage}[b]{0.245\textwidth}
                \centering
        \includegraphics[width=\textwidth]{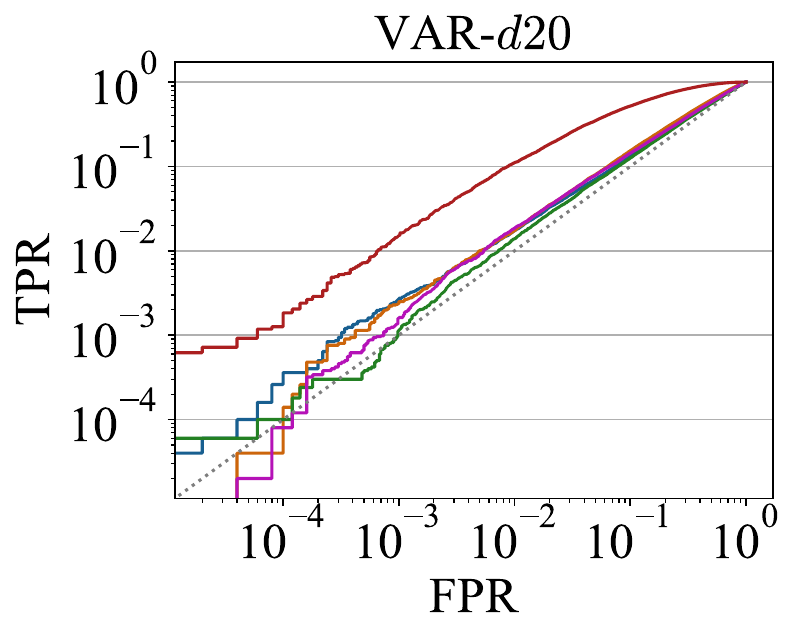}
    \end{minipage}%
    \begin{minipage}[b]{0.245\textwidth}
                \centering
        \includegraphics[width=\textwidth]{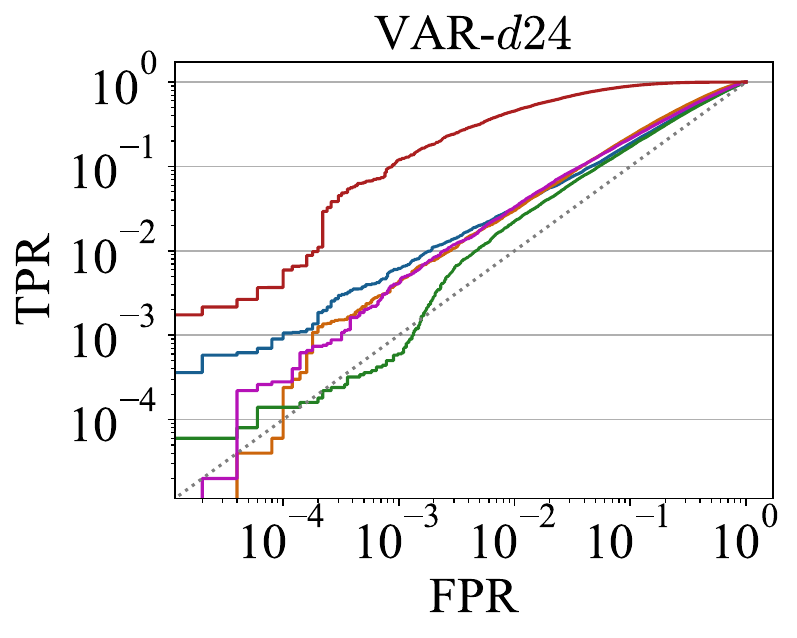}
    \end{minipage}%
    \begin{minipage}[b]{0.245\textwidth}
                \centering
        \includegraphics[width=\textwidth]{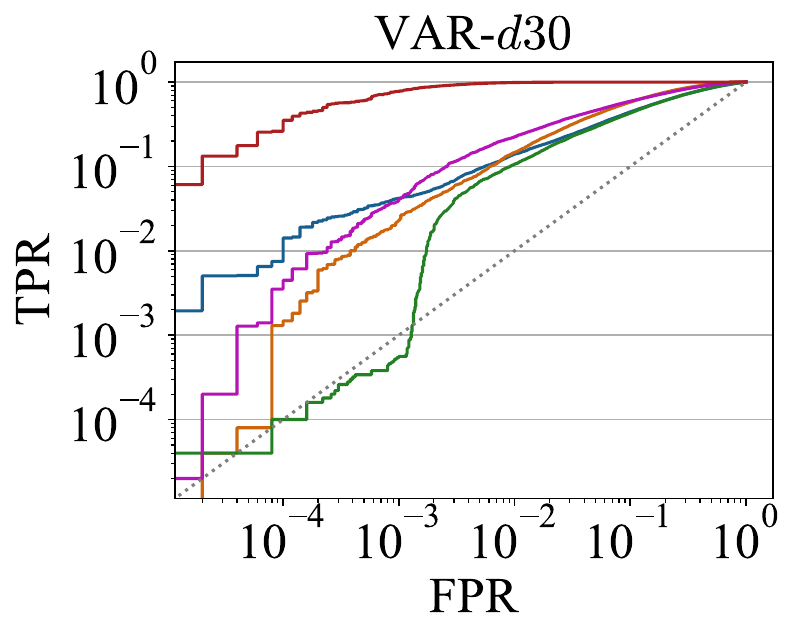}
    \end{minipage}%
    \end{minipage}
    \begin{minipage}[b]{.7\textwidth}
        \centering
        \includegraphics[width=\textwidth]{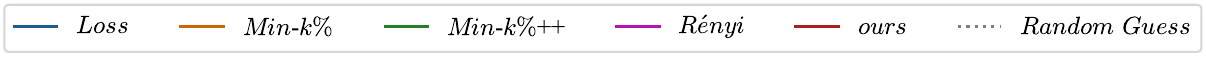}
    \end{minipage}
    \caption{The log ROC curve for membership inference on the VAR models}
    \label{fig: var roc curve}
\end{figure*}

}

\section{Experiments}

\subsection{Experimental Setup.}

\textbf{Target models and datasets.} In the main paper, we utilize three prominent scale-wise visual autoregressive models, including VAR \cite{var}, VAR-CCA \cite{cca}, and VAR-CLIP \cite{varclip}. Inside them, VAR and VAR-CCA are class-conditional generative models, and VAR-CLIP is a text-to-image generative model. Moreover, VAR and VAR-CCA include four different model depths, \ie, $\{d16,d20,d24,d30\}$, and a larger model depth means a larger model. We also conduct experiments on other types of autoregressive models for image generation, such as M-VAR \cite{mvar}, LlamaGen \cite{LlamaGen}, LlamaGen-CCA \cite{cca}, PAR \cite{par}, and NAR \cite{nar}. The results are provided in Appendix B. We conduct experiments on the models without further fine-tuning or other modifications. All the models are trained on the training set of ImageNet \cite{imagenet} with $1000$ classes. 

\textbf{Evaluation Metrics.} Following previous researches \cite{mmmia,mink}, to evaluate our method, we adopt the widely used metrics, including Area-Under-the-ROC-curve (AUROC), and the True Positive Rate when the False Positive Rate is $5\%$ (TPR$@5\%$), and attack success rate (ASR). For ASR, we randomly select a subset of both member and non-member samples, comprising $20\%$ of the total, and compute an optimal threshold from it. Then we apply the threshold to evaluate the ASR for the remaining samples.

\textbf{Baselines.}
Since the autoregressive generation paradigm in visual autoregressive models is similar to the text generation paradigm, we adapt some advanced training data detection methods for large language models as our baseline. 
Specifically, we adapt  the $Loss$ attack \cite{lossattack}, $Min$-$k\%$ \cite{mink}, $Min$-$k\%$++ \cite{minkpp} and \textit{R\'enyi} entropy score \cite{renyi} in our experiments. The implementation detail for baselines is provided in Appendix A.

\textbf{Implementation Details.}
We use the first $50$ images for each class in the training set as the member dataset $\mathbb{D}_{mem}$, and the validation set of ImageNet is supported as the hold-out dataset $\mathbb{D}_{out}$. In total, we have $50000$ member samples and $50000$ non-member samples. For baselines, we carefully adjust their hyperparameters and select the best ones in each setting. For our method, we set $a=1.75$ and $b=1.3$ in the adaptive score aggregation strategy.

\subsection{Comparison to Baselines}

\noindent\textbf{Evaluation on Class-Condition Generation.}
Table \ref{tab:main var} and \ref{tab:main cca} shows the experimental results on the VAR and VAR-CCA models with different model depths, respectively. Compared to the baseline methods, our method consistently outperforms all other approaches across all evaluation metrics and model variants.
Particularly, when the model depth is small (\eg, $16$ and $20$), the baseline algorithms exhibit minimal effectiveness, performing only slightly better than random guessing. In contrast, our method demonstrates strong distinguishability between member and non-member samples.  Notably, it achieve $0.1472$ and $0.2589$ increase on AUROC for VAR models with depths $16$ and $20$, respectively. As the model depth increases, such as in the case of VAR-$d30$, the AUROC scores exceed $0.9980$, indicating that the effective detection performance for larger models. These results clearly show that our method is highly effective in distinguishing training data from non-training data across all variants of the VAR model.

The log ROC curve for detection on the VAR models is shown in Figure \ref{fig: var roc curve}. It is evident that baseline methods sometimes yield lower TPR than random guessing, especially when the FPR is low. 
In these cases, the True Positive Rate (TPR) of the baseline methods can be almost indistinguishable from a random guess, indicating their limited ability to correctly classify training data. 
In contrast, our method demonstrates significantly superior performance, consistently achieving a much higher True Positive Rate (TPR), even at low False Positive Rates (FPRs). The ROC curve for our approach clearly shows a marked separation from the baseline methods, illustrating the effectiveness of our method in distinguishing between member and non-member samples. This strong distinguishability is particularly evident across different model depths, highlighting the effectiveness and reliability of our approach.

\noindent\textbf{Evaluation on Text-to-Image Generation.}
Table \ref{tab:main varclip} presents the membership inference results on the VAR-CLIP model. In this scenario, our detection method achieves an AUROC of $0.8817$, significantly outperforming the baseline methods. Moreover, when compared to the class-conditional generation scenario (\ie, VAR), the AUROC of our method shows an impressive improvement of $0.2$. This substantial gap highlights a key observation: the text generation model exhibits a strong overfitting characteristic to the training text, which makes the model particularly vulnerable to membership inference. This vulnerability is a direct consequence of the model's reliance on the specific training data, further reinforcing the effectiveness of our method in detecting training data in the context of text-to-image generative models.

{
\def\subwidth{.27\textwidth}
\def\legendwidth{.15\textwidth}
\begin{figure*}[htp]
    \centering
    \begin{minipage}{\textwidth}
        \centering
        \begin{minipage}{\subwidth}
            \centering
            \includegraphics[width=\linewidth]{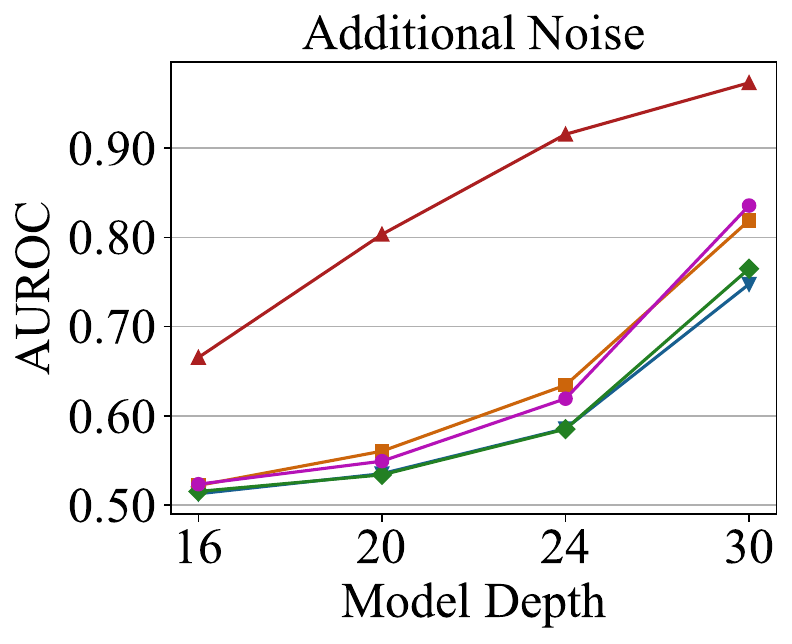}
        \end{minipage}
        \begin{minipage}{\subwidth}
            \centering
            \includegraphics[width=\linewidth]{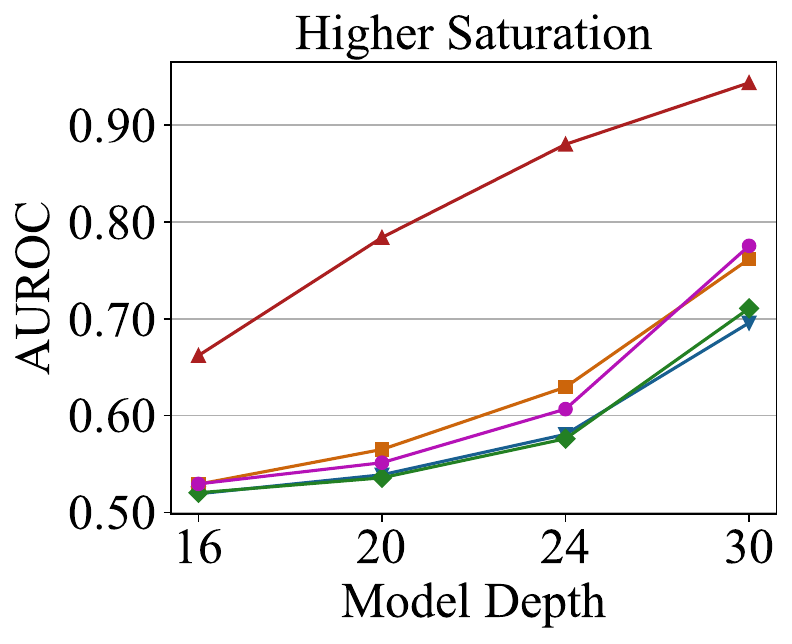}
        \end{minipage}
        \begin{minipage}{\subwidth}
            \centering
            \includegraphics[width=\linewidth]{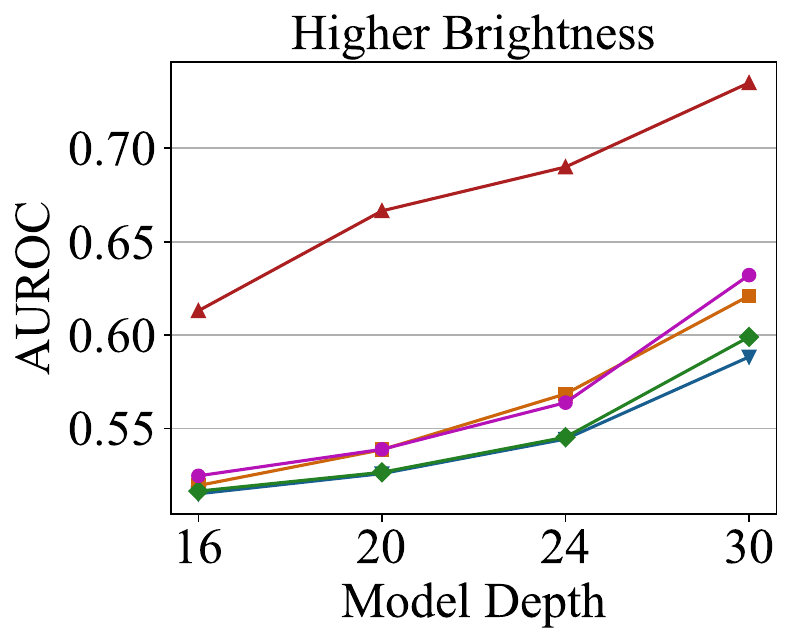}
        \end{minipage}


        \begin{minipage}{\subwidth}
            \centering
            \includegraphics[width=\linewidth]{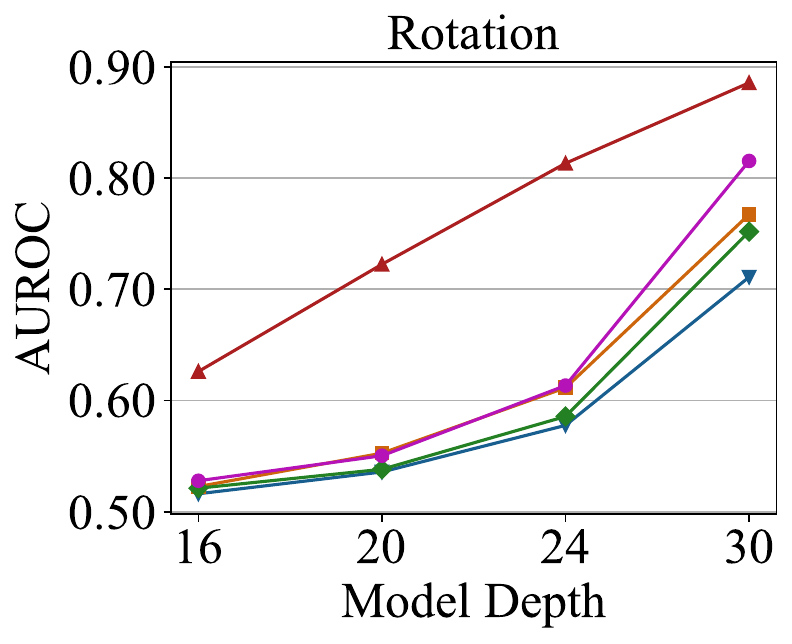}
        \end{minipage}
        \begin{minipage}{\subwidth}
            \centering
            \includegraphics[width=\linewidth]{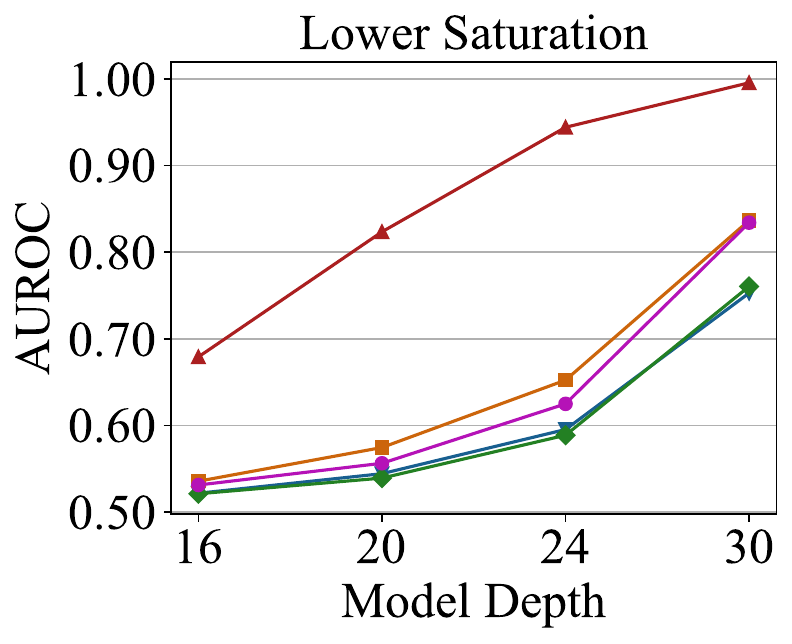}
        \end{minipage}
        \begin{minipage}{\subwidth}
            \centering
            \includegraphics[width=\linewidth]{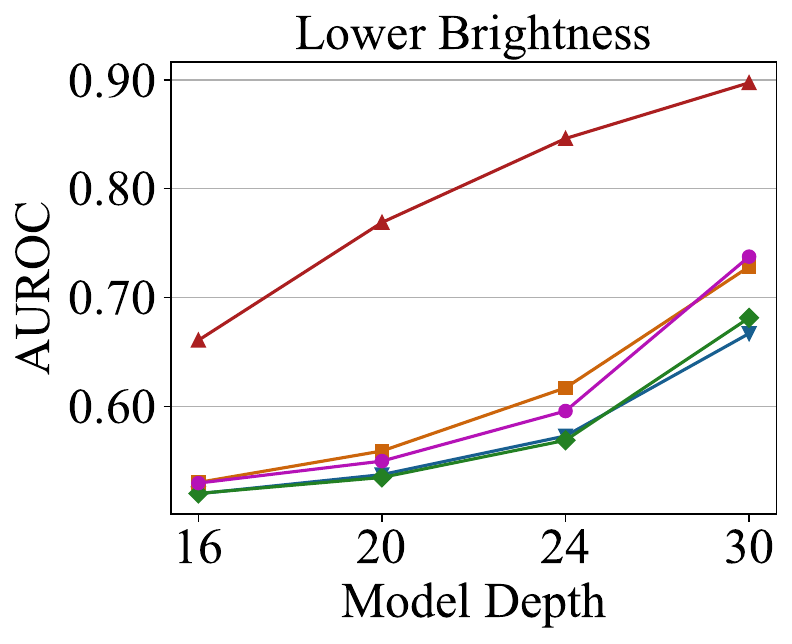}
        \end{minipage}
    \end{minipage}

    \begin{minipage}{.6\textwidth}
            \centering
            \includegraphics[width=\linewidth]{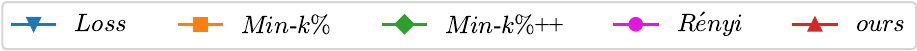}
        \end{minipage}
    \caption{Experimental results on different image transformations}
    \label{fig: aug}
\end{figure*}

}

\begin{table}[!ht]
    \setlength{\tabcolsep}{5pt}
    \normalsize
    \centering
    \caption{Experiment results in VAR-CLIP}
    \label{tab:main varclip}
    \begin{threeparttable} 
    \resizebox{\linewidth}{!}{
    \begin{tabular}{cccc}
        \toprule
 \multirow{2}{*}{{Method}} 
 & \multicolumn{3}{c}{VAR-CLIP} \\

 \cmidrule(lr){2-4} & $\uparrow$ {AUROC} & $\uparrow$ {TPR@5\%}  & $\uparrow$ {ASR}   \\ \midrule

$Loss$ & 0.5537 & 0.0776 & 0.5370  \\
$Min$-$k\%$  & 0.6109 & 0.0894  & 0.5784 \\
$Min$-$k\%$++  & 0.5356 & 0.0624  & 0.5246 \\
\textit{R\'enyi} & 0.5549 & 0.0716  & 0.5384 \\
$ICAS (ours)$ & \textbf{0.8817} & \textbf{0.3989 } & \textbf{0.8024} \\
\bottomrule
    \end{tabular}
    }
    \end{threeparttable}
    
\end{table}

\subsection{Robustness Evaluation}

In real-world scenarios, images are often subjected to various disturbances such as rotation, noise, and other transformations. These factors introduce a certain offset between the original images and the member images used for detection. As a result, it becomes essential to evaluate the robustness of the membership inference under image perturbations to understand its effectiveness in practical settings.
Similar to previous work \cite{mmmia}, we apply four common image transformations to disrupt the original images and test the robustness of our membership inferences:

\begin{itemize}
    \item \textbf{Additional Noise. } Gaussian noise with a standard deviation of $0.1$ is added to the original images, introducing random noise and making the image less clear. Note that the noise is fixed when applying to different images.
    \item \textbf{Rotation.} Images are rotated by $10$ degrees around the geometric midpoint.
    \item \textbf{Saturation.} The saturation levels of images are increased to $150\%$ or decreased to $50\%$, affecting the color intensity and making the image either more vivid or dull.
    \item \textbf{Brightness.} The brightness of images are increased or decreased by $50\%$, affecting the overall lightness or darkness of the images.
\end{itemize}

These perturbations simulate real-world scenarios in which images may undergo various transformations, helping to assess how robust our method are in the presence of such challenges. The results are shown in Figure \ref{fig: aug}. We observe that the AUROC values of different detection algorithms experience varying degrees of decline due to these transformations. Since such perturbations are not part of the training process for the VAR model, transformed member images are not used during training, increasing the difficulty for the detection methods.

However, our proposed method demonstrates exceptional resilience against various types of image transformations, while baseline methods exhibit poor robustness. The experimental results highlight the superior stability and robustness of our approach when confronted with diverse distortions encountered in real-world settings. Specifically, our method maintains a high level of accuracy despite the presence of noise, rotation, and changes in brightness or saturation. This underscores not only the robustness of our method but also its potential applicability in dynamic environments where image transformations are commonplace. Such resilience ensures that our approach remains effective even in challenging, noisy, or altered conditions, setting it apart from baseline methods that struggle to adapt to these real-world distortions.

\subsection{Ablation Study}

This part provides ablation studies to analyze the role of each component in our detection method. We conduct the experiments using VAR with different model depths.

\begin{table}[!ht]
    \setlength{\tabcolsep}{5pt}
    \normalsize
    \centering
    \caption{Ablation study of the implicit classifier (IC) in VAR}
    \label{tab:ablation imc}
    \begin{threeparttable} 
    \resizebox{\linewidth}{!}{
    \begin{tabular}{ccccccc}
        \toprule
 \multirow{2}{*}{{Model}} 
 & \multicolumn{2}{c}{$\uparrow$ AUROC} & \multicolumn{2}{c}{$\uparrow$ TPR@5\%} & \multicolumn{2}{c}{$\uparrow$ ASR} \\

 \cmidrule(lr){2-3}\cmidrule(lr){4-5}\cmidrule(lr){6-7} & \textbf{w} IC & \textbf{w/o} IC &  \textbf{w} IC & \textbf{w/o} IC&  \textbf{w} IC & \textbf{w/o} IC   \\ \midrule

VAR-$d16$ & \textbf{0.6838} & 0.5463 & \textbf{0.1417} & 0.0618 &  \textbf{0.6345} &  0.5340  \\
VAR-$d20$  & \textbf{0.8402} & 0.6031 & \textbf{0.3494} & 0.0889  &  \textbf{0.7628} &  0.5748  \\
VAR-$d24$  & \textbf{0.9624} & 0.6868 & \textbf{0.7824} & 0.1368  &  \textbf{0.9001} &  0.6355  \\
VAR-$d30$  & \textbf{0.9990} & 0.8266 & \textbf{0.9997} & 0.3076  &  \textbf{0.9897} &  0.7505  \\
 
\bottomrule
    \end{tabular}
    }
    \end{threeparttable}
    
\end{table}

\noindent\textbf{Implicit Classifier.}
The implicit classifier plays a crucial role in our detection. In Table \ref{tab:ablation imc}, we present the results of an ablation study that evaluates the impact of removing the implicit classifier. The results show a significant drop in detection performance when the implicit classifier is removed, emphasizing its importance in the overall effectiveness of our approach. This finding reinforces the fact that the implicit classifier plays a vital role in enhancing the discriminative power in the training data detection.

\begin{table}[!ht]
    \setlength{\tabcolsep}{5pt}
    \normalsize
    \centering
    \caption{Ablation study of the Adaptive Score Aggregation Strategy (AS)  in VAR}
    \label{tab:ablation weight}
    \begin{threeparttable} 
    \resizebox{\linewidth}{!}{
    \begin{tabular}{ccccccc}
        \toprule
\multirow{2}{*}{{Model}} 
 & \multicolumn{2}{c}{$\uparrow$ AUROC} & \multicolumn{2}{c}{$\uparrow$ TPR@5\%} & \multicolumn{2}{c}{$\uparrow$ ASR} \\

 \cmidrule(lr){2-3}\cmidrule(lr){4-5}\cmidrule(lr){6-7} & \textbf{w} AS & \textbf{w/o} AS &  \textbf{w} AS & \textbf{w/o} AS &  \textbf{w} AS & \textbf{w/o} AS  \\ \midrule

VAR-$d16$ & \textbf{0.6838} & 0.6354 & \textbf{0.1417} & 0.0856  &  \textbf{0.6345} & 0.6001 \\
VAR-$d20$  & \textbf{0.8402} & 0.7793 & \textbf{0.3494} & 0.1826  &  \textbf{0.7628} & 0.7133  \\
VAR-$d24$  & \textbf{0.9624} & 0.9303 & \textbf{0.7824} & 0.5643   &  \textbf{0.9001} & 0.8613 \\
VAR-$d30$  & \textbf{0.9990} & 0.9955 & \textbf{0.9997} & 0.9969   &  \textbf{0.9897} & 0.9787 \\

\bottomrule
    \end{tabular}
    }
    \end{threeparttable}
    
\end{table}

\begin{figure}[!htbp]
    \begin{subfigure}[b]{\linewidth}
    \begin{minipage}[b]{\textwidth}
                \centering
        \includegraphics[width=\textwidth]{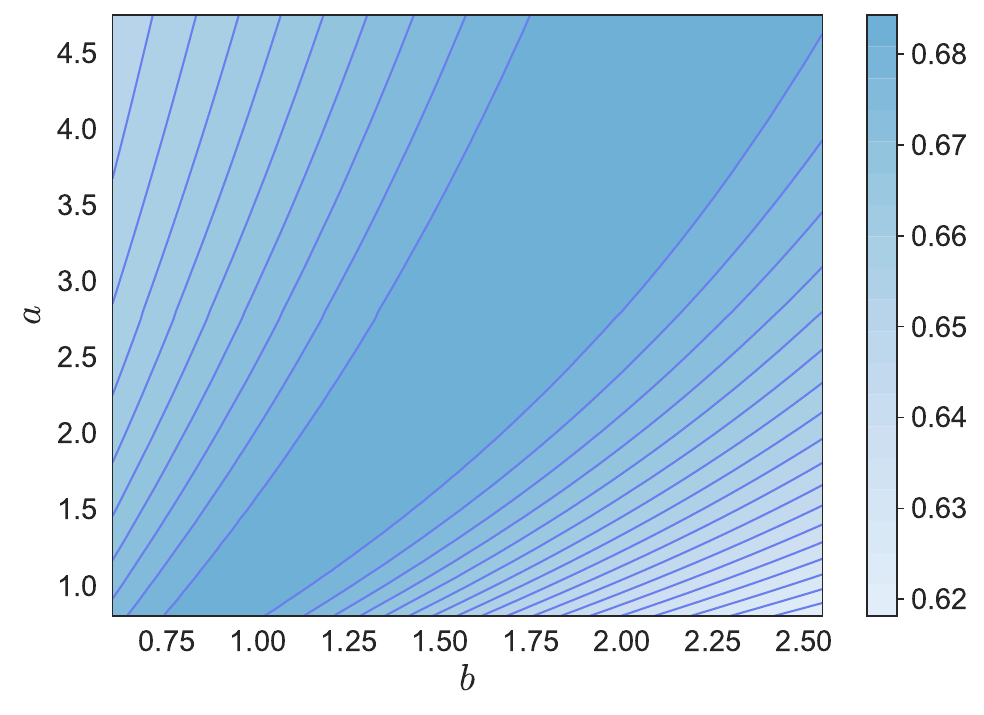}
    \end{minipage}%
    \end{subfigure}
    \caption{AUROC score with different hyperparameters}
    \label{fig: hyperparameters}
\end{figure}

\noindent\textbf{Adaptive Score Aggregation Strategy.}
We conduct an ablation study on the weighted aggregation strategy. By observing the results presented in Table \ref{tab:ablation weight}, we find that removing the weight adjustment strategy leads to a substantial decline in detection performance. This underscores the importance of the weight aggregation approach in our method, highlighting its contribution to improving the performance of the membership inference.

To gain further insight into the impact of this strategy's hyperparameters, we examine the AUROC scores under different parameter settings, as shown in Figure \ref{fig: hyperparameters}. The results indicate that the AUROC score remains relatively stable within a certain range of hyperparameters. Optional hyperparameters are $a=1.75$ and $b=1.3$, and we use them in our main experiments.


\begin{figure}[!htbp]
    \begin{subfigure}[b]{\linewidth}
    \begin{minipage}[b]{\textwidth}
                \centering
        \includegraphics[width=\textwidth]{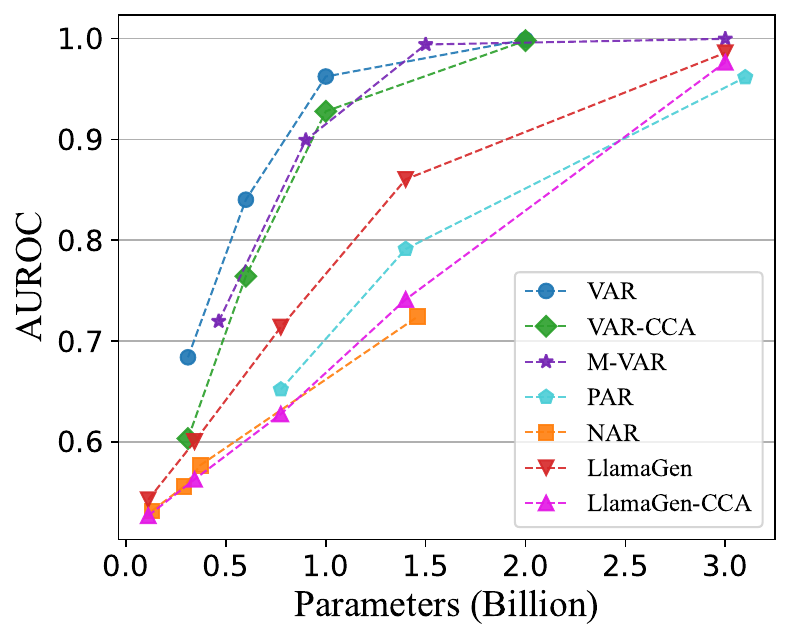}
    \end{minipage}%
    \end{subfigure}
    \caption{AUROC score with different model size}
    \label{fig: scaling law}
\end{figure}

\subsection{MI Linear Scaling Law}

Previous research has established that scaling up autoregressive models such as LLMs \cite{gpt4,scalinglaw1,scalinglaw2} and VAR \cite{var} follows a power-law scaling law like $L=(\beta X)^\alpha$, where $X$ is controllable factors such as model parameters or training tokens, $L$ is the evaluation result like test loss \cite{gpt4} or token error rate \cite{var}.

To explore whether similar scaling laws exist for membership inference, we extend our experiments to various paradigms of autoregressive image generative models with different model sizes, ranging from $ 111 M$ to $ 3.1 B$ parameters. The models under consideration include VAR \cite{var}, VAR-CCA \cite{cca}, M-VAR \cite{mvar}, PAR \cite{par}, NAR \cite{nar}, LlamaGen \cite{LlamaGen}, and LlamaGen-CCA \cite{cca}. We apply our membership inference method to these autoregressive models of varying sizes, and the AUROC scores are presented in Figure \ref{fig: scaling law}.
We can observe that when AUROC is not close to $1$, \eg, below $0.98$, it has a linear scaling law with the model parameters:
\begin{equation}
    \mathrm{AUROC}=\alpha P+\beta,
\end{equation}
where $P$ is the number of model parameters in billions. For example:
\begin{equation}
\begin{aligned}
    \mathrm{AUROC}_{\text{NAR}}&=0.143P+0.517 & r&=0.998 \\
    \mathrm{AUROC}_{\text{LlamaGen-CCA}}&=0.157P+0.511 & r&=0.999,
\end{aligned}
\end{equation}
where $r$ represents the correlation coefficient. This linear scaling law highlights the vulnerability of large autoregressive generative models to membership inference. 

Compared to other autoregressive image generative paradigms, we discover that it is easier to detect the training data on the scale-wise visual autoregressive models (\ie, VAR, VAR-CCA, and M-VAR), given the higher AUROC score. This significant phenomenon can be attributed to how these models process and generate visual data across multiple scales. The granular generation process in scale-wise models may increase the likelihood of memorizing specific details from the training set, thereby making them more susceptible to membership inference.

\section{Conclusion}

In this paper, we explore the training data detection on autoregressive image generative models for the first time. To detect whether an image is used to train the generative model, we first leverage an implicit classifier to calculate scores for all image tokens. Then we propose an adaptive score aggregation strategy to aggregate all token scores into a final score.
The member samples tend to achieve higher final scores. 
Experimental results show that our method outperformed baselines designed in the text autoregressive generative domains.
Robustness and ablation studies demonstrate the robustness and effectiveness of our method. Moreover, we discover a linear scaling law on membership inference and find the significant vulnerability of scale-wise visual autoregressive models.




\bibliographystyle{ACM-Reference-Format}

\clearpage
\appendix

\section{Implement of Baselines}

Since the autoregressive paradigm between visual autoregressive models and large language models is similar, we adopt some membership inference attacks designed for large language models as our baselines.

\subsection{Loss Attack}

The $Loss$ attack method proposed by \cite{lossattack} predicts the membership of a sample using the training loss of the target model. In the training process, the cross-entropy (CE) loss will be used to train the model. To adapt it into autoregressive image generative models, we set the score of a single image token $x_i$ as:
\begin{equation}
    \mathrm{Score}(x_i, c)=\log p_\theta(x_i|c),
\end{equation}
where $c$ is the condition, and $p_\theta$ is the prediction probability from the target model.
The final score is the sum of token scores:
\begin{equation}
    \mathrm{Score}(\textbf{x}, c) = \sum_{i=1}^N  \mathrm{Score}(x_i, c),
\end{equation}
where $N$ is the number of image tokens.

\subsection{Min-k\%}

The token score of the $Min$-$k\%$ method \cite{mink} is the same as the $Loss$ attack. Then, it selects the $k\%$ tokens with the lowest score, denoted as $\textbf{S}_{min}$, where $k$ is a hyperparameter. The final score is defined as the sum of the scores from the selected tokens:
\begin{equation}
    \mathrm{Score}(\textbf{x}, c) = \sum_{i\in \textbf{S}_{min}}  \mathrm{Score}(x_i, c).
\end{equation}

\subsection{Min-k\%++}

The $Min$-$k\%$ method \cite{minkpp} is an upgrade from $Min$-$k\%$. It considers the expected probability of all tokens within the vocabulary. The token score is:
\begin{equation}
    \mathrm{Score}(x_i, c)=\frac{\log p_\theta(x_i|c)-\mu_{\cdot|c}}{\sigma_{\cdot|c}},
\end{equation}
where the $\mu_{\cdot|c}$ and $\sigma_{\cdot|c}$ are the expectation and the standard deviation of the tokens' log probability over the vocabulary. The formulation is as follows:
\begin{equation}
\begin{aligned}
    \mu_{\cdot|c}&=\mathbb{E}_{x\sim p_\theta (\cdot | c)}[\log p_\theta(x|c)] \\
    \sigma_{\cdot|c}&=\sqrt{\mathbb{E}_{x\sim p_\theta (\cdot | c)}[(\log p_\theta(x|c)-\mu_{\cdot|c}])^2}.
\end{aligned}
\end{equation}

The aggregation of the token scores is the same as the $Min$-$k\%$ method.

\subsection{R\'enyi}

This method \cite{renyi} calculates the token scores leveraging the \textit{R\'enyi} entropy \cite{renyientropy} of order $\alpha$ (a hyperparameter):
\begin{equation}
    \mathrm{Score}(x_i, c)=\frac{1}{1-\alpha}\log \sum_{x\in V}(p^\alpha_\theta(x|c)),
\end{equation}
where $V$ is the vocabulary. When $\alpha=1$, the token score is defined as:
\begin{equation}
    \mathrm{Score}(x_i, c)=-\sum_{x\in V}p(x|c)\log p(x|c).
\end{equation}
And when $\alpha=+\infty$, they can be calculated as:
\begin{equation}
    \mathrm{Score}(x_i, c)=-\log \max_{x\in V} p(x|c).
\end{equation}

Then, the method selects the $k\%$ tokens with the highest token scores, denoted as $\textbf{S}_{max}$, where $k$ is a hyperparameter. The final score is defined as the sum of the scores from the selected tokens:
\begin{equation}
    \mathrm{Score}(\textbf{x}, c) = \sum_{i\in \textbf{S}_{max}}  \mathrm{Score}(x_i, c).
\end{equation}

\begin{table*}[h]
    \setlength{\tabcolsep}{5pt}
    \normalsize
    \centering
    \caption{Experiment results in M-VAR models (Part 1).}
    \label{tab:main mvar 1}
    \begin{threeparttable} 
    \resizebox{.7\linewidth}{!}{
    \begin{tabular}{ccccccc}
        \toprule
 \multirow{2}{*}{{Method}} 
 & \multicolumn{3}{c}{M-VAR-$d16$} & \multicolumn{3}{c}{M-VAR-$d20$} \\

 \cmidrule(lr){2-4}\cmidrule(lr){5-7}& $\uparrow$ {AUROC} & $\uparrow$ {TPR@5\%} & $\uparrow$ ASR  & $\uparrow$ {AUROC} & $\uparrow$ {TPR@5\%} & $\uparrow$ ASR  \\ \midrule

$Loss$  & 0.5274 & 0.0594 & 0.5201 & 0.5644 & 0.0797 & 0.5462  \\
$Min$-$k\%$  & 0.5471 & 0.0683 & 0.5319 & 0.6109 & 0.0963 & 0.5795  \\
$Min$-$k\%$++  & 0.5232 & 0.0581 & 0.5147 & 0.5547 & 0.0714 & 0.5355  \\
\textit{R\'enyi}  & 0.5333 & 0.0669 & 0.5241 & 0.5792 & 0.0916 & 0.5555  \\
$ICAS (ours)$  & \textbf{0.7196} & \textbf{0.1792} & \textbf{0.6627} & \textbf{0.8993} & \textbf{0.5132} & \textbf{0.8202}  \\

\bottomrule
    \end{tabular}
    }
    \end{threeparttable}
    
\end{table*}

\begin{table*}[h]
    \setlength{\tabcolsep}{5pt}
    \normalsize
    \centering
    \caption{Experiment results in M-VAR models (Part 2).}
    \label{tab:main mvar 2}
    \begin{threeparttable} 
    \resizebox{.7\linewidth}{!}{
    \begin{tabular}{ccccccc}
        \toprule
 \multirow{2}{*}{{Method}} 
 & \multicolumn{3}{c}{M-VAR-$d24$} & \multicolumn{3}{c}{M-VAR-$d32$} \\

 \cmidrule(lr){2-4}\cmidrule(lr){5-7}& $\uparrow$ {AUROC} & $\uparrow$ {TPR@5\%} & $\uparrow$ ASR  & $\uparrow$ {AUROC} & $\uparrow$ {TPR@5\%} & $\uparrow$ ASR  \\ \midrule

$Loss$  & 0.6605 & 0.1500 & 0.6190 & 0.8181 & 0.3782 & 0.7462  \\
$Min$-$k\%$  & 0.7570 & 0.2081 & 0.6894 & 0.9154 & 0.5643 & 0.8400  \\
$Min$-$k\%$++  & 0.6353 & 0.1170 & 0.5003 & 0.7809 & 0.2849 & 0.5000  \\
\textit{R\'enyi}  & 0.6777 & 0.1649 & 0.6277 & 0.8330 & 0.3997 & 0.7528  \\
$ICAS (ours)$  & \textbf{0.9942} & \textbf{0.9801} & \textbf{0.9658} & \textbf{0.9998} & \textbf{1.0000} & \textbf{0.9967}  \\

\bottomrule
    \end{tabular}
    }
    \end{threeparttable}
    
\end{table*}
\begin{table*}[h]
    \setlength{\tabcolsep}{5pt}
    \normalsize
    \centering
    \caption{Experiment results in PAR models.}
    \label{tab:main par}
    \begin{threeparttable} 
    \resizebox{.95\linewidth}{!}{
    \begin{tabular}{cccccccccc}
        \toprule
 \multirow{2}{*}{{Method}} 
 & \multicolumn{3}{c}{PAR-XL} & \multicolumn{3}{c}{PAR-XXL}& \multicolumn{3}{c}{PAR-3B} \\

 \cmidrule(lr){2-4}\cmidrule(lr){5-7}\cmidrule(lr){8-10} & $\uparrow$ {AUROC} & $\uparrow$ {TPR@5\%} & $\uparrow$ ASR  & $\uparrow$ {AUROC} & $\uparrow$ {TPR@5\%} & $\uparrow$ ASR  & $\uparrow$ {AUROC} & $\uparrow$ {TPR@5\%} & $\uparrow$ ASR \\ \midrule

$Loss$  & 0.5330 & 0.0662 & 0.5263 & 0.5904 & 0.1021 & 0.5689 & 0.7345 & 0.2299 & 0.6799  \\
$Min$-$k\%$  & 0.5370 & 0.0590 & 0.5263 & 0.5997 & 0.0829 & 0.5766 & 0.7478 & 0.2204 & 0.6932  \\
$Min$-$k\%$++  & 0.5275 & 0.0561 & 0.5198 & 0.5941 & 0.0782 & 0.5665 & 0.7566 & 0.2288 & 0.5056  \\
\textit{R\'enyi}  & 0.5389 & 0.0622 & 0.5287 & 0.5927 & 0.0828 & 0.5689 & 0.7327 & 0.2052 & 0.6775  \\
$ICAS (ours)$  & \textbf{0.6519} & \textbf{0.1680} & \textbf{0.6086} & \textbf{0.7913} & \textbf{0.3682} & \textbf{0.7184} & \textbf{0.9616} & \textbf{0.8147} & \textbf{0.8952}  \\

\bottomrule
    \end{tabular}
    }
    \end{threeparttable}
    
\end{table*}

\begin{table*}[h]
    \setlength{\tabcolsep}{5pt}
    \normalsize
    \centering
    \caption{Experiment results in NAR models (Part 1).}
    \label{tab:main nar 1}
    \begin{threeparttable} 
    \resizebox{.7\linewidth}{!}{
    \begin{tabular}{ccccccc}
        \toprule
 \multirow{2}{*}{{Method}} 
 & \multicolumn{3}{c}{NAR-B} & \multicolumn{3}{c}{NAR-M} \\

 \cmidrule(lr){2-4}\cmidrule(lr){5-7}& $\uparrow$ {AUROC} & $\uparrow$ {TPR@5\%} & $\uparrow$ ASR  & $\uparrow$ {AUROC} & $\uparrow$ {TPR@5\%} & $\uparrow$ ASR \\ \midrule

$Loss$  & 0.5055 & 0.0520 & 0.5016 & 0.5099 & 0.0528 & 0.5087\\
$Min$-$k\%$  & 0.5058 & 0.0532 & 0.5046 & 0.5109 & 0.0548 & 0.5078 \\
$Min$-$k\%$++  & 0.5045 & 0.0536 & 0.5041 & 0.5061 & 0.0554 & 0.5053 \\
\textit{R\'enyi}  & 0.5317 & 0.0606 & 0.5210 & 0.5267 & 0.0619 & 0.5193\\
$ICAS (ours)$  & \textbf{0.5315} & \textbf{0.060} & \textbf{0.5210} & \textbf{0.5556} & \textbf{0.0705} & \textbf{0.5386} \\

\bottomrule
    \end{tabular}
    }
    \end{threeparttable}
    
\end{table*}

\begin{table*}[h]
    \setlength{\tabcolsep}{5pt}
    \normalsize
    \centering
    \caption{Experiment results in NAR models (Part 2).}
    \label{tab:main nar 2}
    \begin{threeparttable} 
    \resizebox{.7\linewidth}{!}{
    \begin{tabular}{ccccccc}
        \toprule
 \multirow{2}{*}{{Method}} 
 & \multicolumn{3}{c}{NAR-L} & \multicolumn{3}{c}{NAR-XXL} \\

 \cmidrule(lr){2-4}\cmidrule(lr){5-7}& $\uparrow$ {AUROC} & $\uparrow$ {TPR@5\%} & $\uparrow$ ASR  & $\uparrow$ {AUROC} & $\uparrow$ {TPR@5\%} & $\uparrow$ ASR  \\ \midrule

$Loss$  & 0.5128 & 0.0542 & 0.5064   & 0.5501 & 0.0729 & 0.5359  \\
$Min$-$k\%$  & 0.5166 & 0.0578 & 0.5089  & 0.5637 & 0.0701 & 0.5442  \\
$Min$-$k\%$++  & 0.5103 & 0.0552 & 0.5100  & 0.5320 & 0.0575 & 0.5238  \\
\textit{R\'enyi}  & 0.5268 & 0.0627 & 0.5162  & 0.5463 & 0.0715 & 0.5332  \\
$ICAS (ours)$  & \textbf{0.5768} & \textbf{0.0806} & \textbf{0.5532} & \textbf{0.7241} & \textbf{0.2000} & \textbf{0.6632}  \\

\bottomrule
    \end{tabular}
    }
    \end{threeparttable}
    
\end{table*}

\begin{table*}[h]
    \setlength{\tabcolsep}{5pt}
    \normalsize
    \centering
    \caption{Experiment results in LlamaGen models (Part 1).}
    \label{tab:main llamagen 1}
    \begin{threeparttable} 
    \resizebox{.95\linewidth}{!}{
    \begin{tabular}{cccccccccc}
        \toprule
 \multirow{2}{*}{{Method}} 
 & \multicolumn{3}{c}{LlamaGen-B} & \multicolumn{3}{c}{LlamaGen-L}& \multicolumn{3}{c}{LlamaGen-XL} \\

 \cmidrule(lr){2-4}\cmidrule(lr){5-7}\cmidrule(lr){8-10} & $\uparrow$ {AUROC} & $\uparrow$ {TPR@5\%} & $\uparrow$ ASR  & $\uparrow$ {AUROC} & $\uparrow$ {TPR@5\%} & $\uparrow$ ASR  & $\uparrow$ {AUROC} & $\uparrow$ {TPR@5\%} & $\uparrow$ ASR \\ \midrule

$Loss$  & 0.5047 & 0.0521 & 0.5060 & 0.5170 & 0.0537 & 0.5133 & 0.5404 & 0.0625 & 0.5290  \\
$Min$-$k\%$  & 0.5211 & 0.0579 & 0.5105 & 0.5420 & 0.0651 & 0.5303 & 0.5804 & 0.0814 & 0.5560  \\
$Min$-$k\%$++  & 0.5151 & 0.0564 & 0.5135 & 0.5375 & 0.0652 & 0.5272 & 0.5729 & 0.0826 & 0.5560  \\
\textit{R\'enyi}  & 0.5019 & 0.0475 & 0.5028 & 0.5136 & 0.0498 & 0.5103 & 0.5339 & 0.0595 & 0.5255  \\
$ICAS (ours)$  & \textbf{0.5438} & \textbf{0.0629} & \textbf{0.5323} & \textbf{0.5990} & \textbf{0.0946} & \textbf{0.5698} & \textbf{0.7119} & \textbf{0.1859} & \textbf{0.6529}  \\

\bottomrule
    \end{tabular}
    }
    \end{threeparttable}
    
\end{table*}

\begin{table*}[h]
    \setlength{\tabcolsep}{5pt}
    \normalsize
    \centering
    \caption{Experiment results in LlamaGen models (Part 2).}
    \label{tab:main llamagen 2}
    \begin{threeparttable} 
    \resizebox{.7\linewidth}{!}{
    \begin{tabular}{ccccccc}
        \toprule
 \multirow{2}{*}{{Method}} 
 & \multicolumn{3}{c}{LlamaGen-XXL} & \multicolumn{3}{c}{LlamaGen-3B} \\

 \cmidrule(lr){2-4}\cmidrule(lr){5-7}& $\uparrow$ {AUROC} & $\uparrow$ {TPR@5\%} & $\uparrow$ ASR  & $\uparrow$ {AUROC} & $\uparrow$ {TPR@5\%} & $\uparrow$ ASR  \\ \midrule

$Loss$  & 0.5810 & 0.0806 & 0.5599 & 0.7126 & 0.2054 & 0.6590  \\
$Min$-$k\%$  & 0.6446 & 0.1178 & 0.6046 & 0.8361 & 0.3049 & 0.7603  \\
$Min$-$k\%$++  & 0.6193 & 0.1067 & 0.5890 & 0.7426 & 0.1979 & 0.6870  \\
\textit{R\'enyi}  & 0.5689 & 0.0801 & 0.5497 & 0.6736 & 0.1800 & 0.6313  \\
$ICAS (ours)$  & \textbf{0.8578} & \textbf{0.4372} & \textbf{0.7772} & \textbf{0.9855} & \textbf{0.9317} & \textbf{0.9409}  \\

\bottomrule
    \end{tabular}
    }
    \end{threeparttable}
    
\end{table*}

\begin{table*}[h]
    \setlength{\tabcolsep}{5pt}
    \normalsize
    \centering
    \caption{Experiment results in LlamaGen-CCA models (Part 1).}
    \label{tab:main llamagen cca 1}
    \begin{threeparttable} 
    \resizebox{.95\linewidth}{!}{
    \begin{tabular}{cccccccccc}
        \toprule
 \multirow{2}{*}{{Method}} 
 & \multicolumn{3}{c}{LlamaGen-CCA-B} & \multicolumn{3}{c}{LlamaGen-CCA-L}& \multicolumn{3}{c}{LlamaGen-CCA-XL} \\

 \cmidrule(lr){2-4}\cmidrule(lr){5-7}\cmidrule(lr){8-10} & $\uparrow$ {AUROC} & $\uparrow$ {TPR@5\%} & $\uparrow$ ASR  & $\uparrow$ {AUROC} & $\uparrow$ {TPR@5\%} & $\uparrow$ ASR  & $\uparrow$ {AUROC} & $\uparrow$ {TPR@5\%} & $\uparrow$ ASR \\ \midrule

$Loss$  & 0.5090 & 0.0550 & 0.5076 & 0.5212 & 0.0578 & 0.5153 & 0.5513 & 0.0730 & 0.5348  \\
$Min$-$k\%$  & 0.5255 & 0.0593 & 0.5171 & 0.5503 & 0.0671 & 0.5360 & 0.6038 & 0.0878 & 0.5756  \\
$Min$-$k\%$++  & 0.5187 & 0.0574 & 0.5133 & 0.5457 & 0.0675 & 0.5353 & 0.6011 & 0.0976 & 0.5770  \\
\textit{R\'enyi}  & 0.5047 & 0.0521 & 0.5038 & 0.5164 & 0.0532 & 0.5126 & 0.5406 & 0.0653 & 0.5261  \\
$ICAS (ours)$  & \textbf{0.5267} & \textbf{0.0624} & \textbf{0.5198} & \textbf{0.5627} & \textbf{0.0835} & \textbf{0.5459} & \textbf{0.6274} & \textbf{0.1425} & \textbf{0.5931}  \\

\bottomrule
    \end{tabular}
    }
    \end{threeparttable}
    
\end{table*}

\begin{table*}[h]
    \setlength{\tabcolsep}{5pt}
    \normalsize
    \centering
    \caption{Experiment results in LlamaGen-CCA models (Part 2).}
    \label{tab:main llamagen cca 2}
    \begin{threeparttable} 
    \resizebox{.7\linewidth}{!}{
    \begin{tabular}{ccccccc}
        \toprule
 \multirow{2}{*}{{Method}} 
 & \multicolumn{3}{c}{LlamaGen-CCA-XXL} & \multicolumn{3}{c}{LlamaGen-CCA-3B} \\

 \cmidrule(lr){2-4}\cmidrule(lr){5-7}& $\uparrow$ {AUROC} & $\uparrow$ {TPR@5\%} & $\uparrow$ ASR  & $\uparrow$ {AUROC} & $\uparrow$ {TPR@5\%} & $\uparrow$ ASR  \\ \midrule

$Loss$  & 0.6087 & 0.1097 & 0.5778 & 0.8338 & 0.4999 & 0.7606  \\
$Min$-$k\%$  & 0.7008 & 0.1443 & 0.6483 & 0.9533 & 0.6764 & 0.9044  \\
$Min$-$k\%$++  & 0.6958 & 0.1704 & 0.6455 & 0.9397 & 0.7697 & 0.8849  \\
\textit{R\'enyi}  & 0.5867 & 0.0969 & 0.5624 & 0.7598 & 0.3588 & 0.7013  \\
$ICAS (ours)$  & \textbf{0.7412} & \textbf{0.2826} & \textbf{0.6791} & \textbf{0.9766} & \textbf{0.8827} & \textbf{0.9219}  \\
\bottomrule
    \end{tabular}
    }
    \end{threeparttable}
    
\end{table*}

\section{More Experimental Results}
Instead of the scale-wise visual autoregressive models like VAR and VAR-CCA, we also extend our detection methods to other autoregressive image generative models, including M-VAR (Table \ref{tab:main mvar 1} and \ref{tab:main mvar 2}), PAR (Table \ref{tab:main par}), NAR (Table \ref{tab:main nar 1} and \ref{tab:main nar 2}), LlamaGen (Table \ref{tab:main llamagen 1} and \ref{tab:main llamagen 2}) and LlamaGen-CCA (Table \ref{tab:main llamagen cca 1} and \ref{tab:main llamagen cca 2}). All the results indicate our best performance on detecting training data of autoregressive generative models.

\section{MI Scaling Law on Other Perspectives}

\subsection{The number of scales and token counts}

We explore the linear relationship between the AUROC and the use number of scales $k$ or the log of the token count $L$. The results are presented in the left of Fig. \ref{fig: scales}, and they can be approximately fitted as:
$$
\begin{aligned}
\mathrm{AUROC}&=0.0184k + 0.5122 & r=0.9864\\
\mathrm{AUROC}&=0.0189\log_2 L + 0.5057 & r=0.9980.
\end{aligned}
$$

\subsection{Baseline methods}

To explore whether the baseline methods also satisfies the scaling law, we conduct experiments on VAR with different baselines. The results are shown in the right of Fig. \ref{fig: scales}, and they can be approximately fitted as:

$$
\begin{aligned}
\mathrm{AUROC}_{Loss}&=0.1538P+0.4614&r=0.9946\\
\mathrm{AUROC}_{Min-k\%}&=0.1934P+0.4715&r=0.9995\\
\mathrm{AUROC}_{Min-k\%++}&=0.1571P+0.4547&r=0.9904\\
\mathrm{AUROC}_{\textit{R\'enyi}}&=0.1959P+0.4521&r=0.9939.
\end{aligned}
$$
{

\begin{figure}[!htbp]
    \begin{minipage}[b]{0.22\textwidth}
                \centering
        \includegraphics[width=\textwidth]{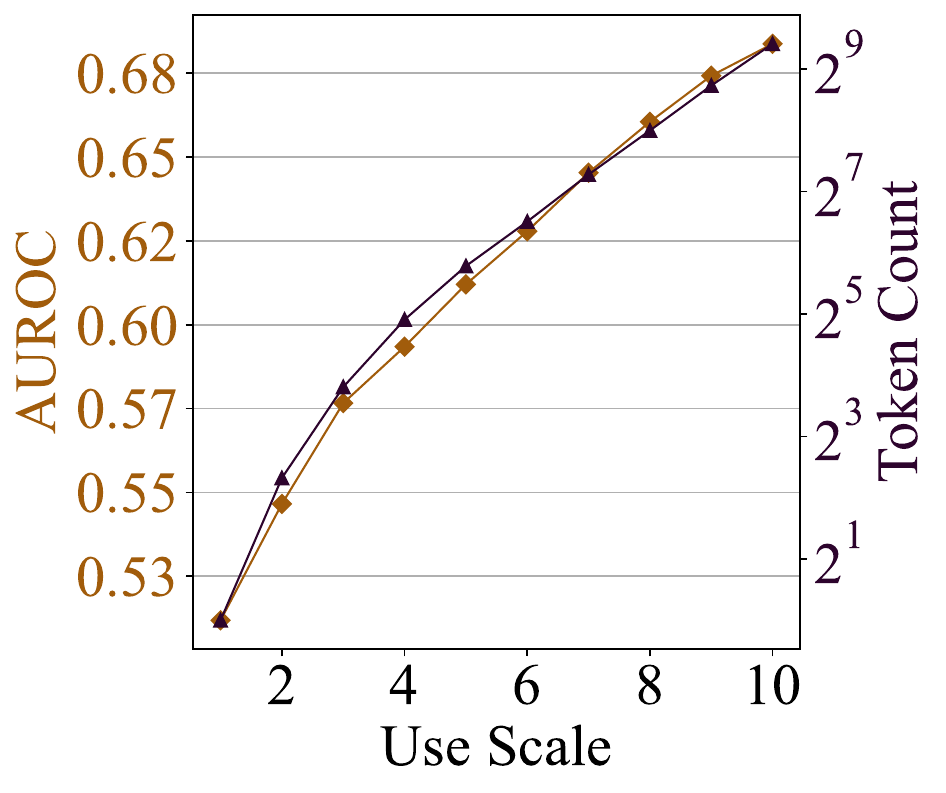}
    \end{minipage}%
    \begin{minipage}[b]{0.22\textwidth}
                \centering
        \includegraphics[width=\textwidth]{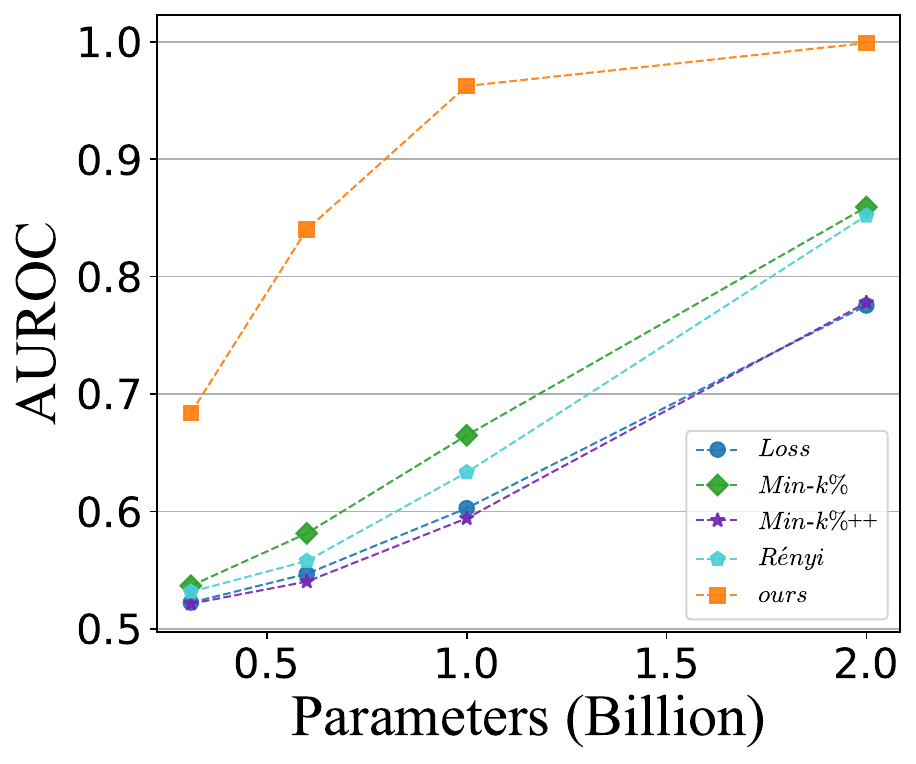}
    \end{minipage}%
    \caption{(\textbf{Left}) The AUROC and token count of VAR-$d16$ when using diffrent number of scales. (\textbf{Right}) The scaling law of AUROC with different methods.}
    \label{fig: scales}
\end{figure}

}
\section{Limitations}

Our method relies on autoregressive generative models that support both conditional and unconditional prediction. As a result, it is not compatible with earlier autoregressive image generation models, such as VQVAE \cite{vqvae} and VQGAN \cite{VQGAN}, which only support conditional prediction.

\section{Future Work}

A promising direction for future research is the extension of our approach to more fine-grained generative paradigms. For example, models like ControlVAR \cite{controlvar} and EditAR \cite{editar} leverage images as conditioning inputs, rather than class labels or textual descriptions. Exploring methods to simulate the unconditional generation probability could also prove to be a valuable avenue for further investigation.

Additionally, autoregressive generation has been expanded to other modalities, such as video \cite{nar} and audio \cite{aar}. Assessing the applicability of our approach to these modalities will be an important direction for future work.

\end{document}